\documentclass[letterpaper, 10 pt, conference]{ieeeconf}  

\IEEEoverridecommandlockouts                              
\usepackage{amsmath} 
\usepackage{amssymb}  
\usepackage{xspace}      
\usepackage{url}         
\usepackage{rotating}
\usepackage{tikz}
\usetikzlibrary{calc,fit}

\graphicspath{%
 {./figures/}
}

\title{\LARGE \bf
Stereo-based terrain traversability analysis using normal-based segmentation and superpixel surface analysis 
}
\author{Aras R. Dargazany
\thanks{The author was with Robotics Research Lab, Department of Computer Science,
        University of Kaiserslautern, Germany}
}

\begin{document}


\maketitle
\thispagestyle{empty}
\pagestyle{empty}

\begin{abstract}
In this paper, an stereo-based traversability analysis approach for all terrains in 
off-road mobile robotics, e.g. Unmanned Ground Vehicles (UGVs) is proposed.
This approach reformulates the problem of terrain traversability analysis into two main problems: 
(1) 3D terrain reconstruction and (2) terrain all surfaces detection and analysis.
The proposed approach is using stereo camera for perception and 3D reconstruction of the terrain.
In order to detect all the existing surfaces in the 3D reconstructed terrain as superpixel surfaces (i.e. segments),
an image segmentation technique is applied using geometry-based features (pixel-based surface normals).
Having detected all the surfaces, Superpixel Surface Traversability Analysis approach (SSTA) is applied 
on all of the detected surfaces (superpixel segments) in order to classify them based on their traversability index.
The proposed SSTA approach is based on: (1) Superpixel surface normal and plane estimation, (2) Traversability analysis using superpixel surface planes. 
Having analyzed all the superpixel surfaces based on their traversability, these surfaces are finally classified into five main categories as following: 
traversable, semi-traversable, non-traversable, unknown and undecided.
\end{abstract}

\section{Introduction}
\label{sec:intro}
The first step for autonomous navigation of UGVs (i.e. Unmanned Ground Vehicles) is the environment perception and 3D terrain reconstruction and 
finding out where it is possible to traverse or which part of the terrain is traversable.
This Analysis of the acquired sensor data in order to distinguish between the traversable and non-traversable areas in the environment 
is known as \textbf{\emph{terrain traversability analysis}}.

\subsection{Motivation}
\begin{figure}[thpb]
  \centering
  \includegraphics[scale=2]{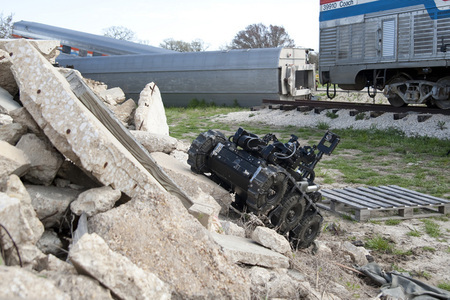}
  \caption{Navigation of small UGV (Unmanned Ground Vehicle) in disaster city in 2010.}
  \label{fig:intro:disaster}
\end{figure}

In ICARUS project (i.e Integrated Components for Assisted Search and Unmanned Rescue) \cite{icarus}, 
two UGVs (i.e Unmanned Ground Vehicles) are employed for coping with disastrous terrains 
such as earthquakes, collapsed buildings and etc.
These two UGVs are composed of a large and a small one which are employed in the above scenarios as shown in figure ~\ref{fig:intro:disaster}.
Autonomous navigation is essential due to likelihood of losing signal for controlling the UGVs using tele-operation in the above scenarios.
That is why terrain traversability analysis is the first step for autonomous navigation of these UGVs.

\subsection{Contribution} 
The proposed approach is mainly reformulating the problem of terrain traversability analysis into four main problems:
  \begin{itemize}
   \item Environment perception and 3D reconstruction using stereo camera
   \item Feature extraction using geometry-based (pixel-based surface normals)
   \item Segmentation of terrain for all surfaces detection - conversion of terrain into collection of superpixel surfaces
   \item Terrain classification - SSTF for classification of all detected surfaces based on traversability
  \end{itemize}

\subsection{Outline}
The remaining paper is structured as follows:
\begin{itemize}
 \item Section \ref{sec:sota} reviews the similar approaches to the proposed approach.
 \item Section \ref{sec:prop} explains the proposed approach in details.
 \item Section \ref{sec:res} shows the experimental results.
 \item Section \ref{sec:disc} concludes our work and explain our ideas for future.
\end{itemize}

\section{Related work}
\label{sec:sota}

Having reviewed the literature in terrain traversability analysis, most of the existing approaches use geometry-based features for 
describing the geometry of the terrain, accurate analysis and estimation of terrain traversability for safe navigation of UGVs.

Our recent work \cite{Dargazany14} in terrain traversability estimation mainly proposed geometry based features such as pixel-based surface normals
using a stereo camera for environment perception.
Mainly, it was explained how these pixel-based surface normals can perform the generated terrain point cloud segmentation 
and terrain classification based on traversability criteria
although the classification results were not accurate and robust enough in some cases such as:
\begin{itemize}
 \item low quality of the generated point cloud
 \item lack of the dominant ground plane
 \item false classification of segments based on the traversability criteria such as max traversable step and slope
 \item step analysis was not clear enough or ignored
\end{itemize}

In \cite{Trevor12} and \cite{trevor2013efficient}, the similar segmentation technique is proposed for detecting drivable road surfaces using geometry-based features 
such as pixel-based normals, euclidean distance using RGB-D cameras such as Kinect and stereo camera.
This approach is applied to planar road surface such as roads which does not work robustly and accurately for rough off-road terrain.

In \cite{bogoslavskyi2013efficient}, a similar approach has been proposed for terrain traversability analysis using Kinect on mobile robots.
This approach is mainly based on geometry-based pixel-based surface normals and considering the kinematic capability of the vehicle 
such as max height, max slope and max step.
This approach was applied to the very rough and disastrous terrain and produced reasonable results using only Kinect
which makes it hard to know how it works with stereo camera and in offline outdoor terrain
since the point cloud generated by Kinect is much cleaner and denser than the point cloud generated by stereo camera.

In \cite{bellone2013unevenness} and \cite{bellone20143d}, an geometry-based feature 
is proposed for roughness estimation so-called as Unevenness Point Descriptor (UPD). 
This feature is basically describing the unevenness and roughness on one point by measuring pixel-based normals and averaging the normals in k-neighborhood.
This proposed feature is describing surface unevenness based on the average surface normals of the points in one neighborhood. 
This approach is not robust enough in very rough environment since this approach is not 
including the kinematics capability of the different UGVs in traversability analysis.

\section{Proposed approach}
\label{sec:prop}
\begin{figure}
  \centering
  \includegraphics[width=0.8\columnwidth]{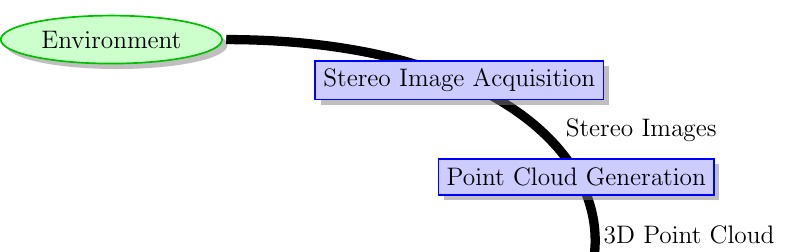}
  \caption{The proposed approach stereo-based 3D terrain reconstruction is shown.}
  \label{fig:prop:workflow_stereo}
\end{figure}

\begin{figure}
  \centering
  \includegraphics[width=0.8\columnwidth]{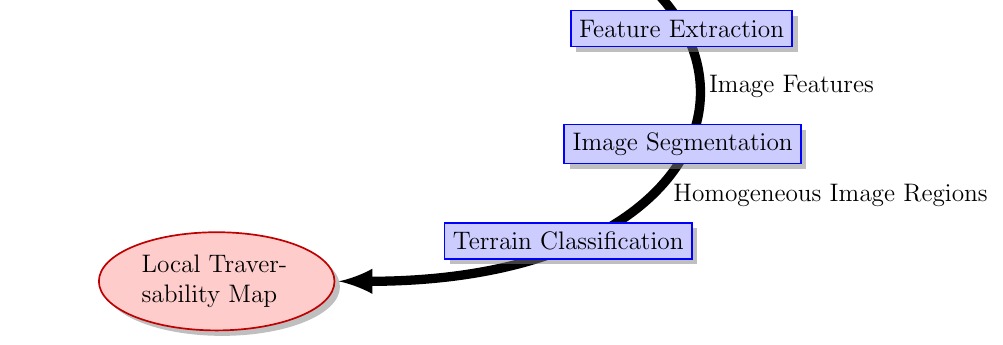}
  \caption{The proposed segmentation-based SSTA (Superpixel Surface Traversability Analysis) is shown.}
  \label{fig:prop:workflow_trav}
\end{figure}

The proposed algorithm can be mainly explained into two main steps: 
\textbf{stereo-based 3D terrain reconstruction} as shown in figure \ref{fig:prop:workflow_stereo} 
and \textbf{segmentation-based SSTA (Superpixel Surface Traversability Analysis)} as shown in figure \ref{fig:prop:workflow_trav}. 

\subsection{Stereo image acquisition}
\begin{figure}
  \centering
  \includegraphics[width=1.0\columnwidth]{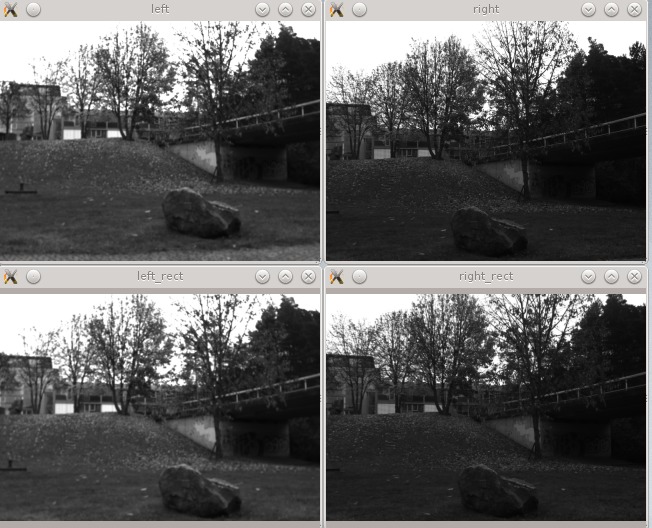}
  \caption{Stereo images are shown before and after rectification: (top) before rectification and (bottom) after rectification and cropped.}
  \label{fig:prop:stereo:calib}
\end{figure}

\begin{figure}
  \centering
  \includegraphics[width=1.0\columnwidth]{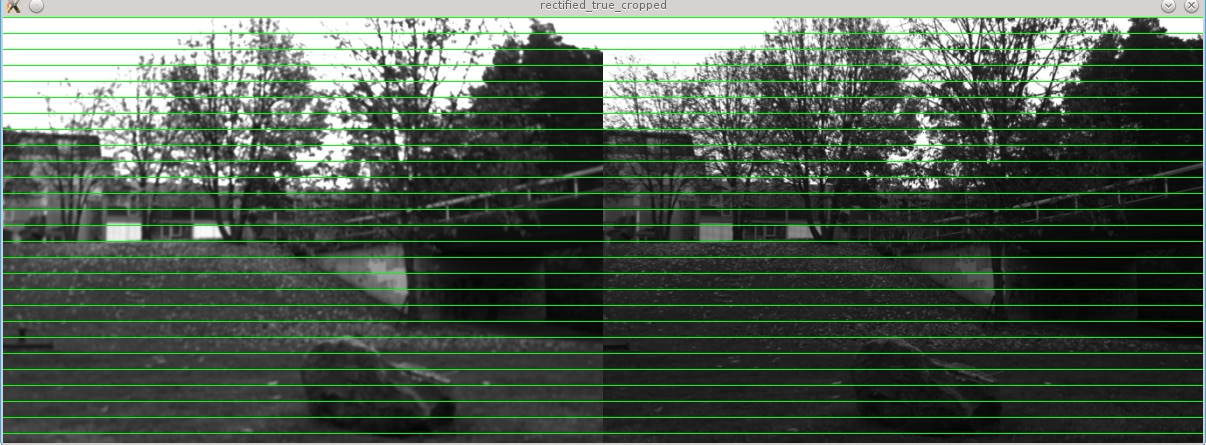}
  \caption{Stereo rectification lines (epipolar lines) are visualized on already rectified and cropped left and right images.}
  \label{fig:prop:stereo:calib_rect}
\end{figure}

Environment perception can be performed using different visual sensors such as stereo camera, RGB-D cameras (Xtion and Kinect) or Time-of-Flight cameras (ToF).
In this work, it is assumed that capturing well-exposed images 
in outdoor settings is used for image acquisition such as Complementary metal–oxide–semiconductor (CMOS) sensors, High-dynamic-range imaging (HDR).
An offline calibration is performed for stereo image rectification using OpenCV stereo calibration tool \cite{opencv}.
The raw images before rectification and after rectification are shown in figure \ref{fig:prop:stereo:calib} and figure \ref{fig:prop:stereo:calib_rect}.
Once stereo images are rectified (shown in figure ~\ref{fig:prop:stereo:calib_rect}) the minimum rectified common area in both of the left and right images 
are measured and used for cropping the images. 

\subsection{Point cloud generation}
The generated point clouds are called organized dense point clouds due to their image-like structure since they are basically generated using images.
Having rectified the stereo images, the corresponding point clouds will be generated as below:

\subsubsection{\textbf{Stereo matching}}
For point cloud generation, stereo matching is applied on stereo images for detecting the corresponding pixels on left and right images
in order to measure the distance between (or displacement of) the corresponding left and right pixel since this distance (disparity) is required  later on for \textbf{triangulation} to measure the depth value.
All of the stereo matching techniques are mainly divided into two main categories of local and global approaches \cite{Tombari10}.
For better evaluation, two stereo matching techniques were chosen for disparity map generation using the already rectified stereo images:
\textbf{Block-based stereo matching (BB)} \cite{bb}, \textbf{Adaptive Cost - 2-Pass Scanline Optimization (ACSO)} \cite{acso}. 

\begin{figure}
  \centering
  \includegraphics[width=1.0\columnwidth]{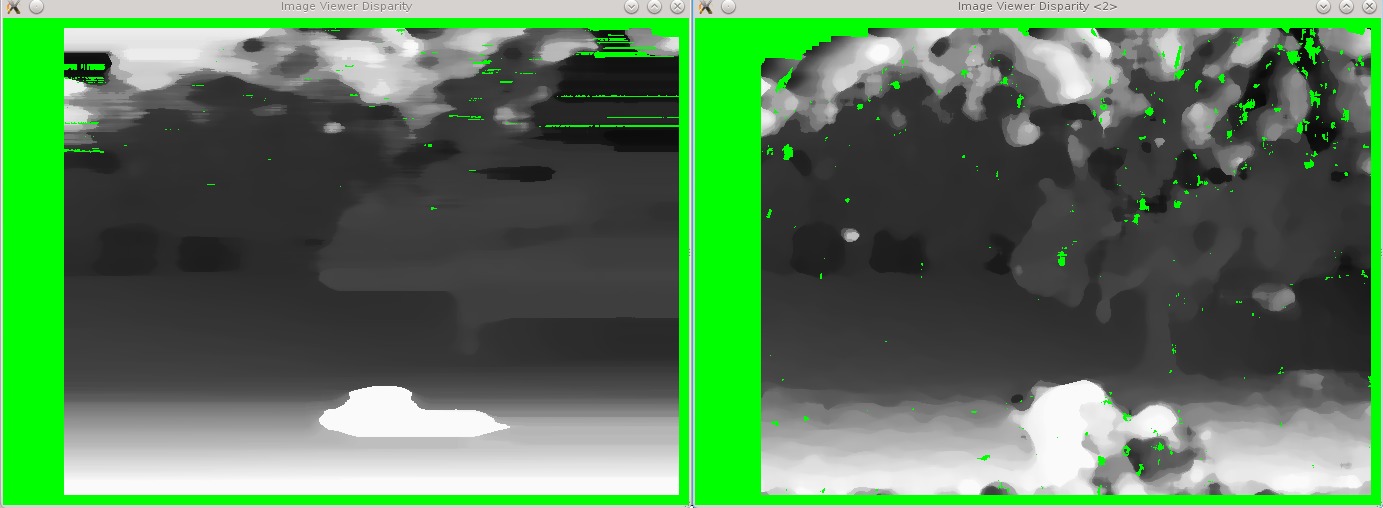}
  \caption{Disparity images generated using: (left) ACSO (Adaptive Cost 2-pass Scanline Optimization) and (right) BB (Block Based)}
  \label{fig:prop:stereo:disp}
\end{figure}

\begin{figure}
 \centering
 \includegraphics[scale=0.23]{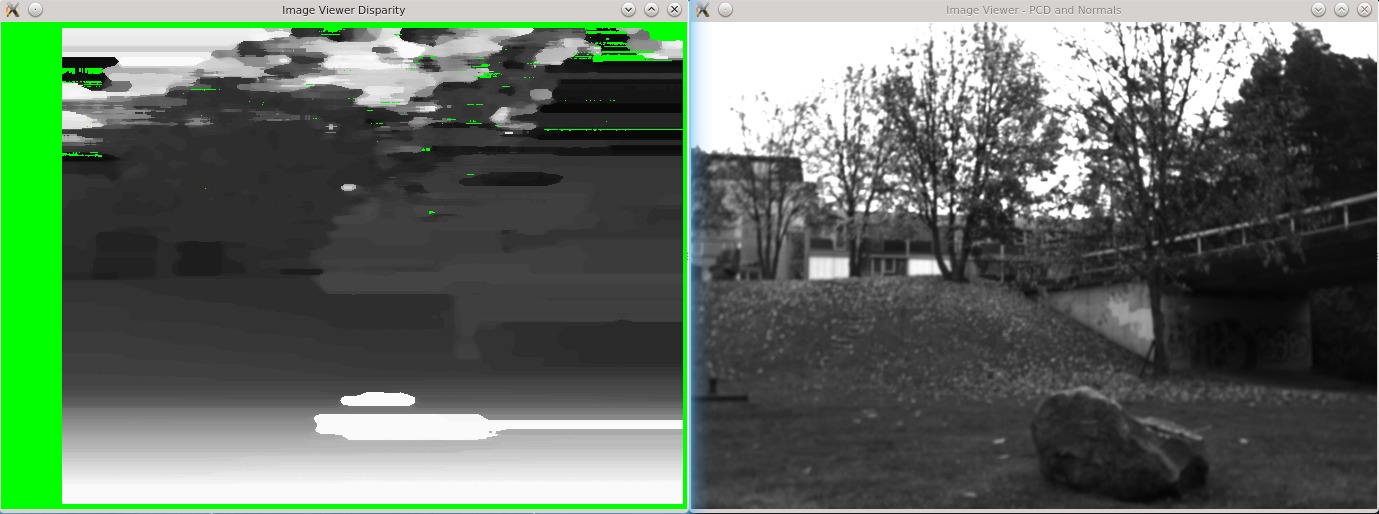}
  \caption{Disparity image generated using ACSO (Adaptive Cost 2-pass Scanline Optimization) vs the original rectified image}
 \label{fig:prop:stereo:disp_acso}
\end{figure}

The yellow area in figure ~\ref{fig:prop:stereo:disp} shows that there is no disparity values available and 
the remaining in gray-scale is showing the disparity values ranging from short distance (white or 255 value) to long distance (black or 0 value).
These two different state-of-the-art approaches are used for generating dense organized point clouds in order to have a better comparison of the traversability results.
As shown in figure \ref{fig:prop:stereo:disp} and also based on experimental, ACSO (Adaptive Cost 2-pass Scanline Optimization) \cite{Tombari10} is much more accurate but slower. 
The generated disparity map using one of these approaches, ACSO (Adaptive Cost 2-pass Scanline Optimization) \cite{Tombari10} illustrated in figure \ref{fig:prop:stereo:disp_acso}.

\subsubsection{\textbf{3D reconstruction}}
Having calculated the disparity map, it is possible to generate the point cloud associated to the left image (or reconstruct the left image in 3D)
using \textbf{triangulation} to measure the depth value.
Left image is chosen to be reconstructed in 3D using the below equations: 
$$
  Z = (focal/disparity) * baseline 
$$
$$ 
 X = ((U-Up)/focal) * Z 
$$
$$
  Y = ((V-Vp)/focal) * Z
$$

\begin{figure}
 \centering
 \includegraphics[width=1.0\columnwidth]{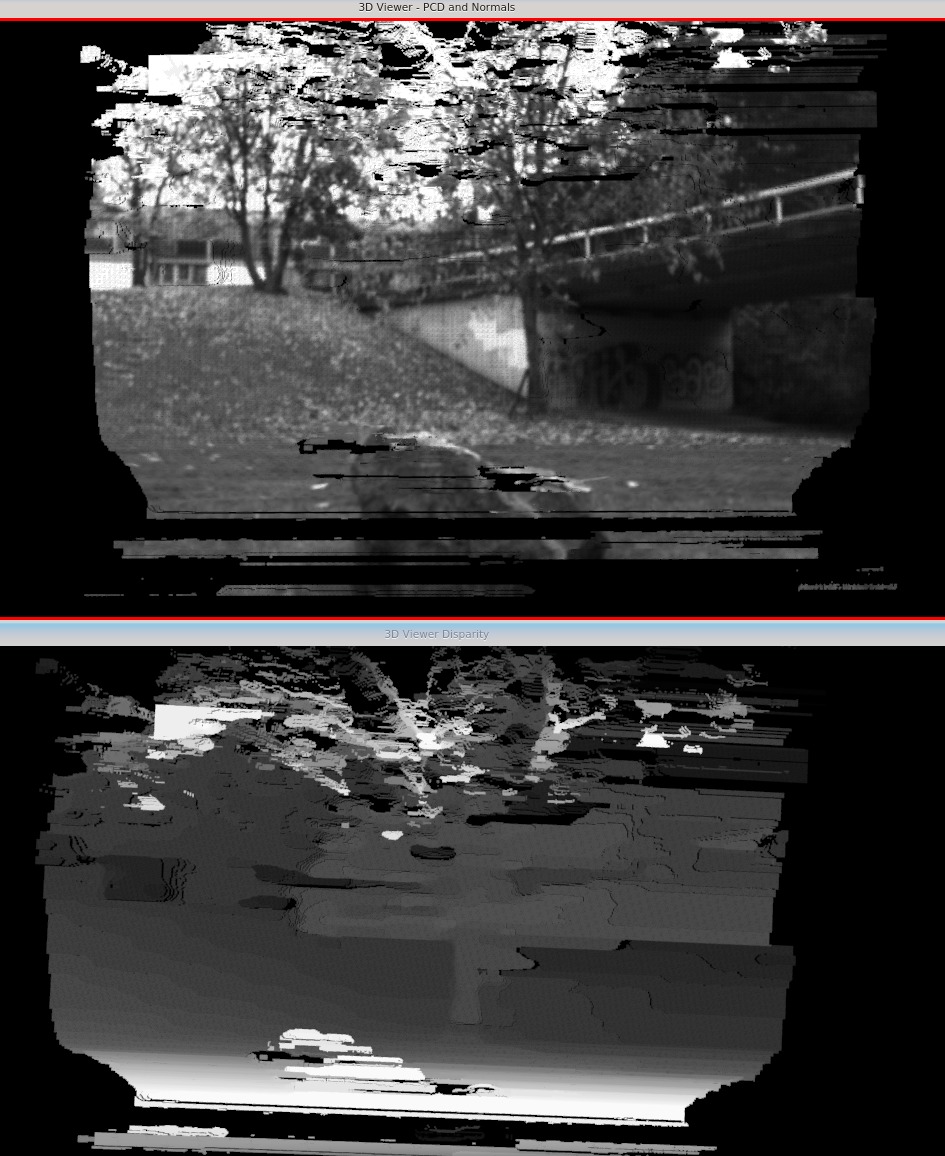}
 \caption{3D visualization of 3D reconstructed: (top) left rectified image and (bottom) resulting disparity image using ACSO approach}
 \label{fig:prop:stereo:pcd}
\end{figure}

In figure \ref{fig:prop:stereo:pcd}, the 3D reconstructed disparity image is showing $(X, Y, Z)$
along with the corresponding disparity value generated by ACSO stereo matching approach.

\subsection{Feature extraction}
Extracting geometry-based features from the generated point cloud of the surrounding environment
is an initial step since these features are required later on for detecting all the superpixel surfaces in the terrain 
which will be performed using terrain segmentation technique. 

Among all of the geometry-based features, two of the most widely used geometric point features, at a query point $p$ on the surface based on its neighboring points, 
are: \textbf{curvatures} and \textbf{normals}. 
Both of these features are considered local since they describe the local neighboring points (i.e. descriptors).
There are three ways of computing normals based on \cite{pcl}.
These three ways are different in optimizing the trade-off between the most accurate normals at every point in a point cloud vs the fast way of computing them as below:
\begin{itemize}
 \item COVARIANCE MATRIX - creates 9 integral images to compute the normal for a specific point from the covariance matrix of its local neighborhood.
 \item AVERAGE 3D GRADIENT - creates 6 integral images to compute smoothed versions of horizontal and vertical 3D gradients and 
 computes the normals using the cross-product between these two gradients.
 \item AVERAGE DEPTH CHANGE - creates only a single integral image and computes the normals from the average depth changes.
\end{itemize}

\begin{figure}
 \centering
\includegraphics[width=1.0\columnwidth]{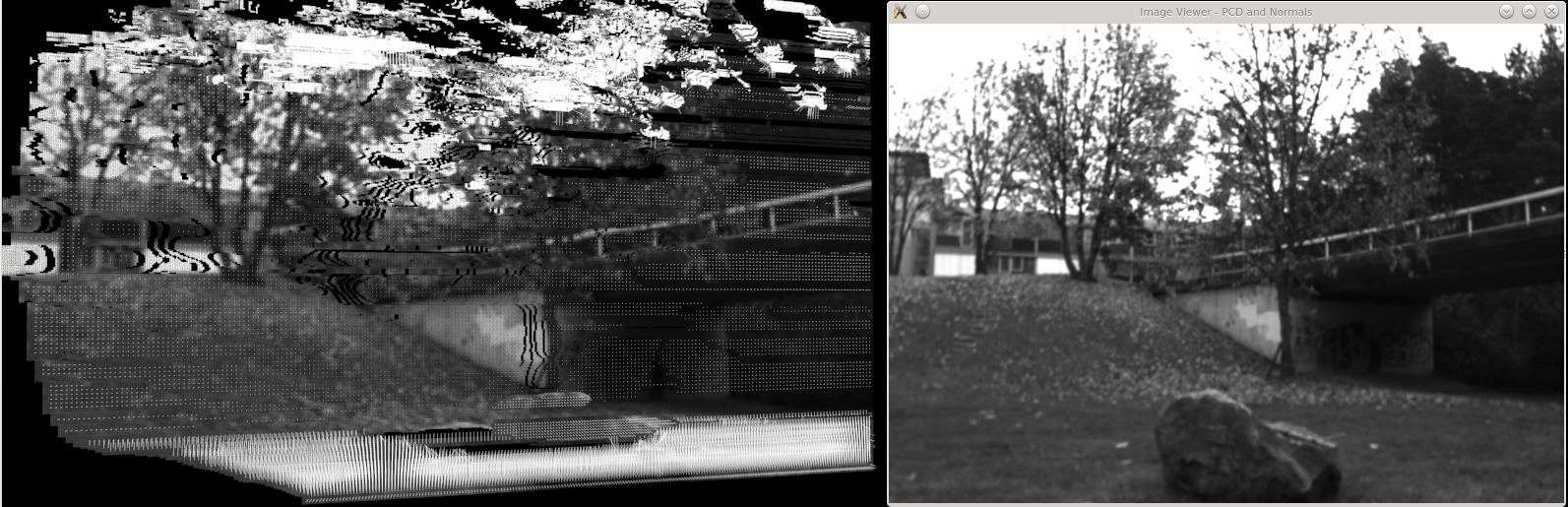}
 \caption{The resulting extracted normals from point cloud is visualized in 3D vs the original rectified image.}
 \label{fig:prop:features:normals:normals_pcd}
\end{figure}

Pixel-based (point-based) normal features are extracted from the generated point cloud using integral images (COVARIANCE MATRIX as mentioned and explained above) \cite{pcl}.
Using these pixel-based normals, it is possible to estimate the roughness of surfaces as shown in figure \ref{fig:prop:features:normals:normals_pcd}.

\subsection{Image segmentation} 
Terrain all surfaces detection is performed by collecting all superpixel surfaces (important ones) using the extracted geometry-based features in last section.
This surface detection is accomplished using the segmentation technique which is basically converting the generated point cloud into a collection of superpixel surfaces (segments).
These superpixel surfaces are basically representing the important surfaces in the 3D reconstructed surrounding environment.

Connected component segmentation technique (as proposed in \cite{Trevor12}) in a 4-connected sense is used as following:
\begin{itemize}
 \item Different feature spaces such as euclidean, surface normals, color and textures: pixel-based (point-based) normal features
 \item Comparison function: the angles between the pixel-based surface normal feature vectors in one segment. 
 \item Neighborhood size: 4-connected sense meaning 4 main directional neighborhood for comparisons such as top, bottom, left and right.
\end{itemize}

\begin{figure}
 \centering
 \includegraphics[scale=0.2]{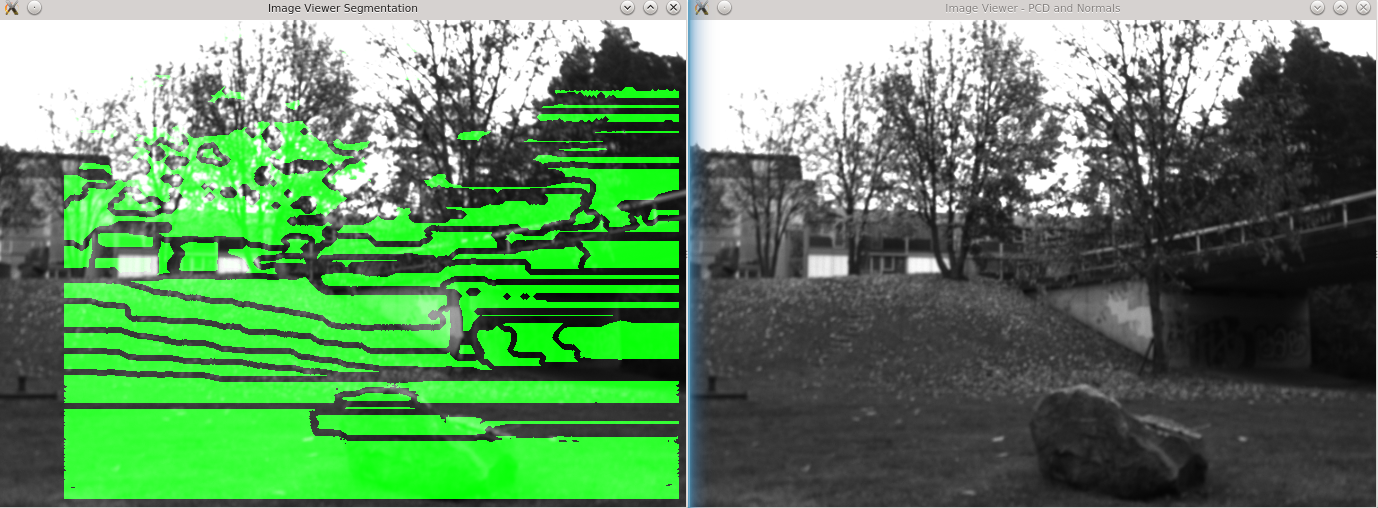}
 \caption{The segmentation result is shown: (left) the original rectified left image, (right) the segmented image.}
 \label{fig:prop:segm:img_segm}
\end{figure}

The segmentation results are shown in figure \ref{fig:prop:segm:img_segm}.

The difference between one point normal vector $n_{p_1}$ and neighboring point normal vector $n_{p_2}$ 
should be smaller than roughness $\alpha_r$ as below:
$$
\vec{n_{p_1}} \bullet \vec{n_{p_2}} \leq \cos{\left(\alpha_r\right)} \eqno{(1)}
$$

\subsection{Terrain classification}
\begin{figure}
 \centering
 \includegraphics[width=1.0\columnwidth]{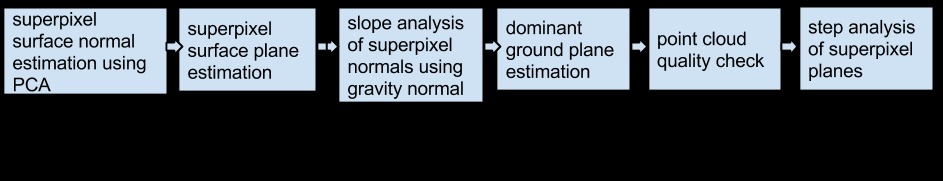}
 \caption{Terrain classification using SSTA (Superpixel Surface Traversability Analysis) is visualized in modular architecture.}
 \label{fig:prop:class:ssa_module} 
\end{figure}

SSTA (Superpixel Surface Traversability Analysis) is applied to all detected surfaces for classification based on their traversability
In this section, the resulting segments (all detected superpixel surfaces) are analyzed based on their point distribution using PCA, approximate each segment with a plane, 
define the required traversability parameters and criteria needed for classification and traversability index generation 
as shown in modular architecture in figure \ref{fig:prop:class:ssa_module}.

These required traversability parameters are included in the proposed SSTA approach:
\begin{itemize}
 \item \textbf{max surface roughness} - required for segmentation (superpixel surface detection) based on the environment
 \item \textbf{gravity normal} $\vec{g_c} = (0, 0, -1)$: this needs to measured in World Coordinate System (WCS) using inertial measurement unit (IMU) and also the tilting angle of the stereo camera.
 \item \textbf{max slope} $alpha_{max}$ - the maximum traversable slope based on UGV kinematic capabilities.
 \item \textbf{max step} $h_{max}$ - the maximum possible step (i.e. max height or gap) the robot can climb related to vehicles mechanical capabilities.
\end{itemize}
These traversability parameters are mainly based on location of the mounted stereo camera, vehicle type, size, kinematics capability
and the application environment or scenario.

\subsubsection{\textbf{Superpixel surface normal estimation using PCA}}
In this part, the point distribution analysis on each superpixel surfaces or segments are performed using PCA.
Given this distribution, the surface plane parameters estimation is measured for each segments.
In order to classify these superpixel surfaces, these segments will be analyzed based on their point distribution as below:
\begin{enumerate}
 \item Minimum inlying points $inliers_{min}$ in each segment - a threshold to avoid noisy segments:
 This threshold is set to a ratio of image height and width 
 as follows $ inliers_{min} = img_{w} \bullet img_{h} \bullet 0.02 \label{trav4}$.
 \item Computing mean and covariance for each segments. 
 \item Applying PCA on the calculated covariance for each segment, analyzing each segments in eigen space and extracting the eigen values and eigen vectors
 \item Using the smallest eigenvector for each segment, the superpixel surface normal is located at centroid of the segment
 (center of gravity on segment) $(A, B, C)$ 
 \item Given centroid (mean) and super pixel surface normal (segment plane normal), it is possible to transform the segment into 
 one superpixel surface using centroid (center of gravity or mean of the segment) and its surface normal
 in order to define the coefficients of the approximated segment plane $Ax+By+Cz+D=0$
\end{enumerate}

\subsubsection{\textbf{Superpixel surface plane estimation}}
Given the superpixel surface normal, it is possible to approximate every segment (superpixel surface) with one plane (planar surface) using its plane parameters such as:
\begin{itemize}
 \item hessian normal $(A, B, C)$ 
 \item offset from origin $(D)$
 \item centroid of the plane (center of gravity) which the computed mean $(X_{c}, Y_{c}, Z_{c})$
\end{itemize}

In order to classify the terrain based its traversability, all estimated superpixel surface planes should be analyzed based on their slope and step.
The detailed description of this analysis is provided in this subsection.

\subsubsection{\textbf{Slope analysis of superpixel normals using gravity normal}}
In order to analyze the slope of the segmented planes in the generated point cloud, gravity in camera coordinate system is required.
This gravity is measured as below equation:
$$
\vec{g_c} = \vec{g} \bullet \cos{(a_w + a_r)}  \eqno{(2)}
$$
The expected gravity normal (or ground plane normal) is $n = (0, 0, -1)$ if the camera is not tilted. 
Usually in mobile robots, IMU sensor are used for more accurate estimation of gravity normal.
In this work, the latter is performed by applying the stereo camera tilting angle $(Theta)$ to 
the Hessian expected ground plane normal $(0, 0, 1)$ as also explained above.

This slope is max traversable ramp or sloped surface, the vehicle or UGV can drive.
As mentioned before, this is based the vehicle information and type.
This parameter is used for slope thresholding of the segments with the nominal ground plane normal.
For traversability estimation of the segments (which are all now approximated using planes), a comparison function is used.
This comparison function is comparing the normals of the segments with gravity vector (nominal ground plane normal) 
using max traversable slope as threshold (degree in our case).

\subsubsection{Slope analysis}
In this part, the plane normals of the segments $\vec{n_p}$ and gravity in the camera coordinate system $\vec{g_c}$ are compared as below equation:
$$%
 \vec{n_p} \bullet \vec{g_c} \leq \cos{\left(\alpha_{max}\right)} \label{eq:slope}
 \vec{n_p} \bullet \vec{g_c} \leq \cos{\left(\alpha_{max}\right)} \eqno{(3)}
$$
Using this slope analysis, a traversability index can be assigned to all the superpixel surface planes: traversable, semi-traversable and non-traversable.

\subsubsection{\textbf{Dominant ground plane estimation}} 
Having classified the segments, it is possible to detect the largest traversable terrain superpixel surface plane (the segment with the max number of inlying points).
This dominant traversable segment is required for:
\begin{enumerate}
 \item Step detection using other segments plane centroid to make sure if they are traversable or not.
 \item Quality of the generated dense point cloud can also be measured based on the confidence level of the dominant traversable superpixel surface plane.
\end{enumerate}

\subsubsection{Dominant ground plane detection and point cloud quality check}
Among all the traversable superpixel surface planes, the dominant ground plane can be detected by:
\begin{itemize}
 \item Counting the number of inlying points $inliers_{min}$. 
This threshold is set using image height and width as follows:
$$ 
inliers_{min} = img_{w} \bullet img_{h} \bullet 0.02 \eqno{(4)}
$$
  \item Superpixel surface normal should be positive.
  \item location of this plane (or height of this plane) should be below the stereo camera or underneath the UGV wheels.
\end{itemize}

\subsubsection{\textbf{Point cloud quality check}}
Having detected the dominant ground plane, it is possible to analyze the quality of the generated point cloud based on the quality of the dominant ground plane
since this plane is the most confident traversable superpixel surface in the generated point cloud 
so that it is highly possible that it indicates the main ground plane on which the vehicle is moving and traversing.
If the dominant traversable segment is not confident enough, it is highly probable that 
the generated point cloud quality is not high and not clean enough for traversability analysis.
Therefore, this point cloud is discarded and skipped to the next generated point cloud.

\subsubsection{\textbf{Step analysis of superpixel planes}}
This is usually a very well-known parameter which is also based on vehicle type.%
This parameters is used for detecting the steps between two segments.
This is max step (also known as gap, height or elevation) between two consecutive planes (or stairs) that a robot can traverse on.
Having detected the dominant traversable segment, it is possible to measure the distance between the centroid of other superpixel surface plane 
to the dominant one using point to plane distance and compare it with max step. This point-to-plane distance (a perpendicular distance) can be measured as follows:
$$
\left| p_d-p_c \right| \leq h_{max} \eqno{(5)} 
$$

\section{Experimental results}
\label{sec:res}
To evaluate the performance of the proposed terrain traversability estimation method, the whole terrain cloud (or image) is classified into five classes:
\begin{enumerate}
 \item \textbf{Traversable} region is \textbf{green}.
 \item \textbf{Semi-traversable} regions are in \textbf{blue}. 
 \item \textbf{Non-traversable obstacles} regions are colorized in \textbf{red}.
 \item \textbf{Unknown} regions are colorized in \textbf{black} - no depth or no disparity values
 \item \textbf{Undecided} no color
 \end{enumerate}
  
\begin{figure}
 \centering
 \includegraphics[width=1.0\columnwidth]{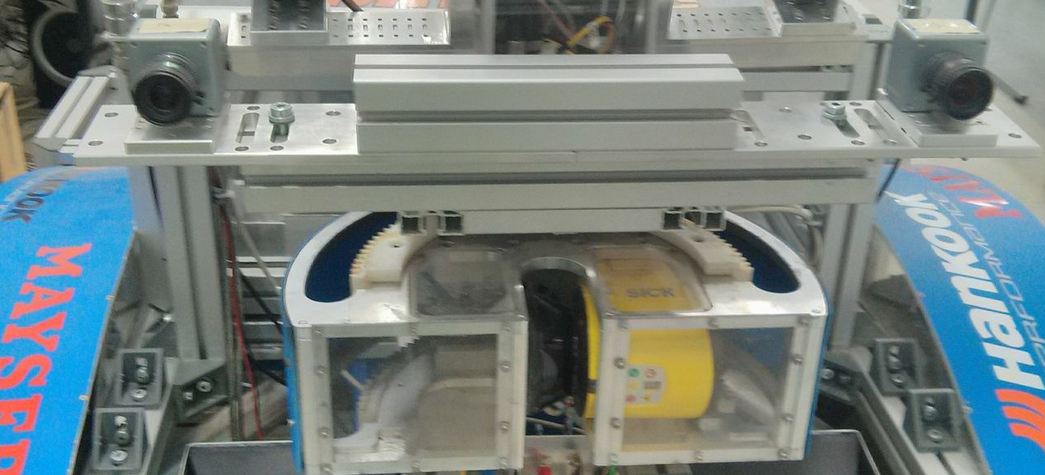}
 \caption{This stereo camera is mounted on platform RAVON \cite{Armbrust10} to generate the gray-scale stereo images in our experiments.}
 \label{fig:res:mine:stereo_ravon_gray} 
\end{figure}

\begin{figure}
 \centering
 \includegraphics[width=1.0\columnwidth]{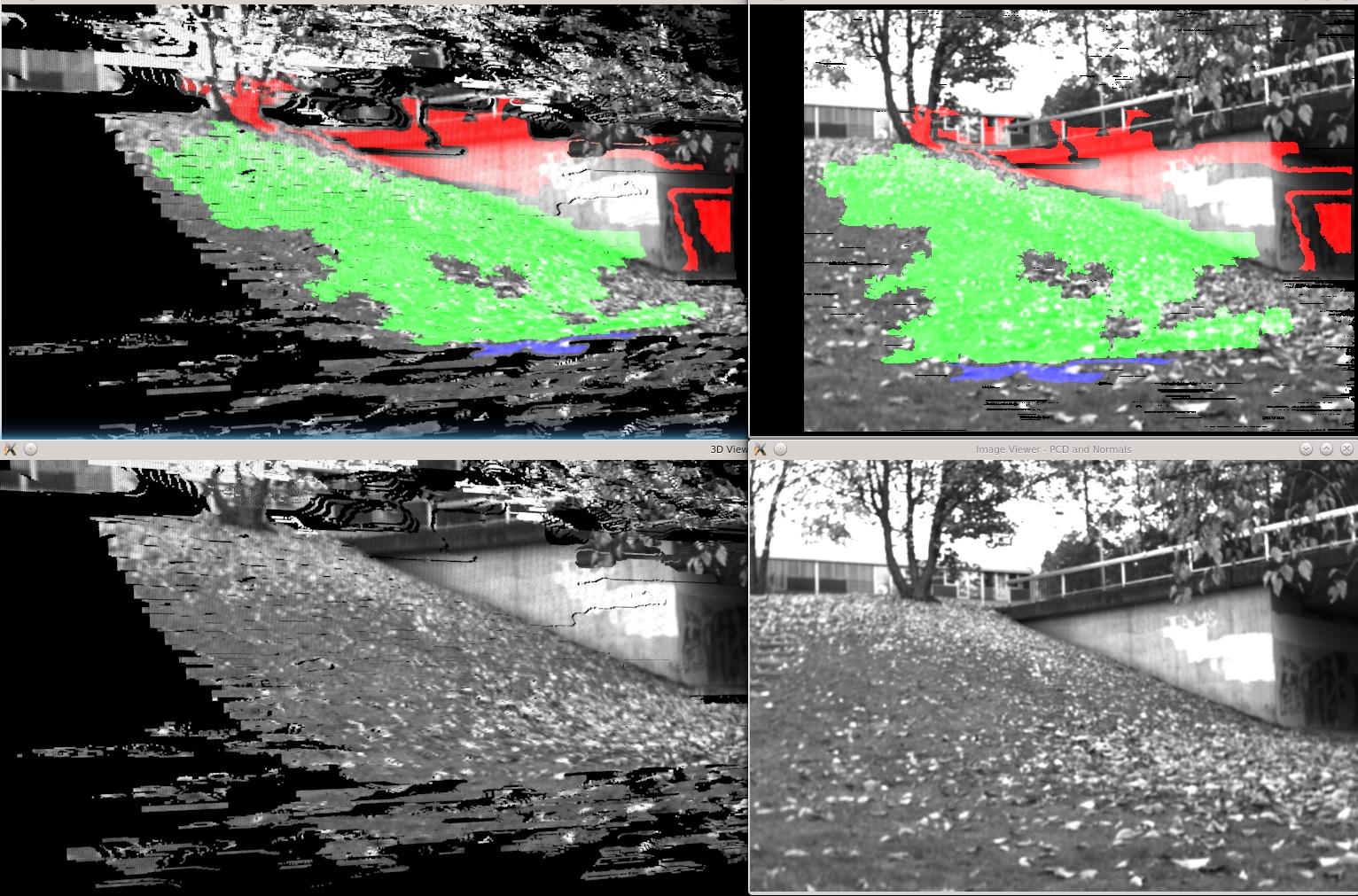}
 \caption{Good frame no. 8: (top row) terrain classification results and (bottom row) 3D reconstructed terrain.}
 \label{fig:res:mine:trav_vs_input8}
\end{figure}

\begin{figure}
 \centering
 \includegraphics[width=1.0\columnwidth]{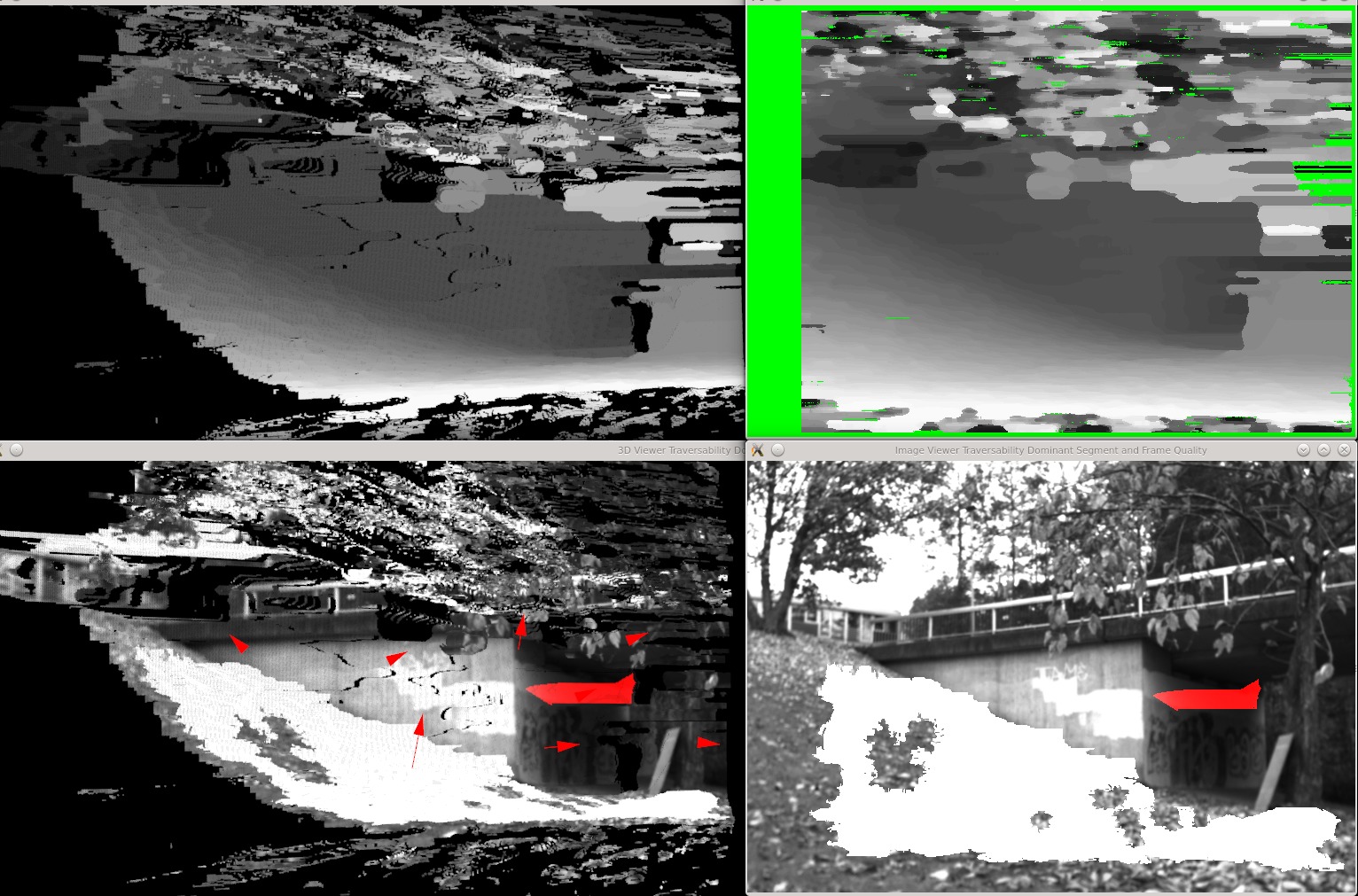}
 \caption{Good frame no. 8: (top row) the disparity image and (bottom row) the dominant ground plane is shown in white and red indicates the dominant obstacle plane. 
 The arrows on the 3D point cloud in left hand is indicating the detected superpixel surface normals.}
 \label{fig:res:mine:disp_vs_domin9}
\end{figure}

\begin{figure}
 \centering
 \includegraphics[width=1.0\columnwidth]{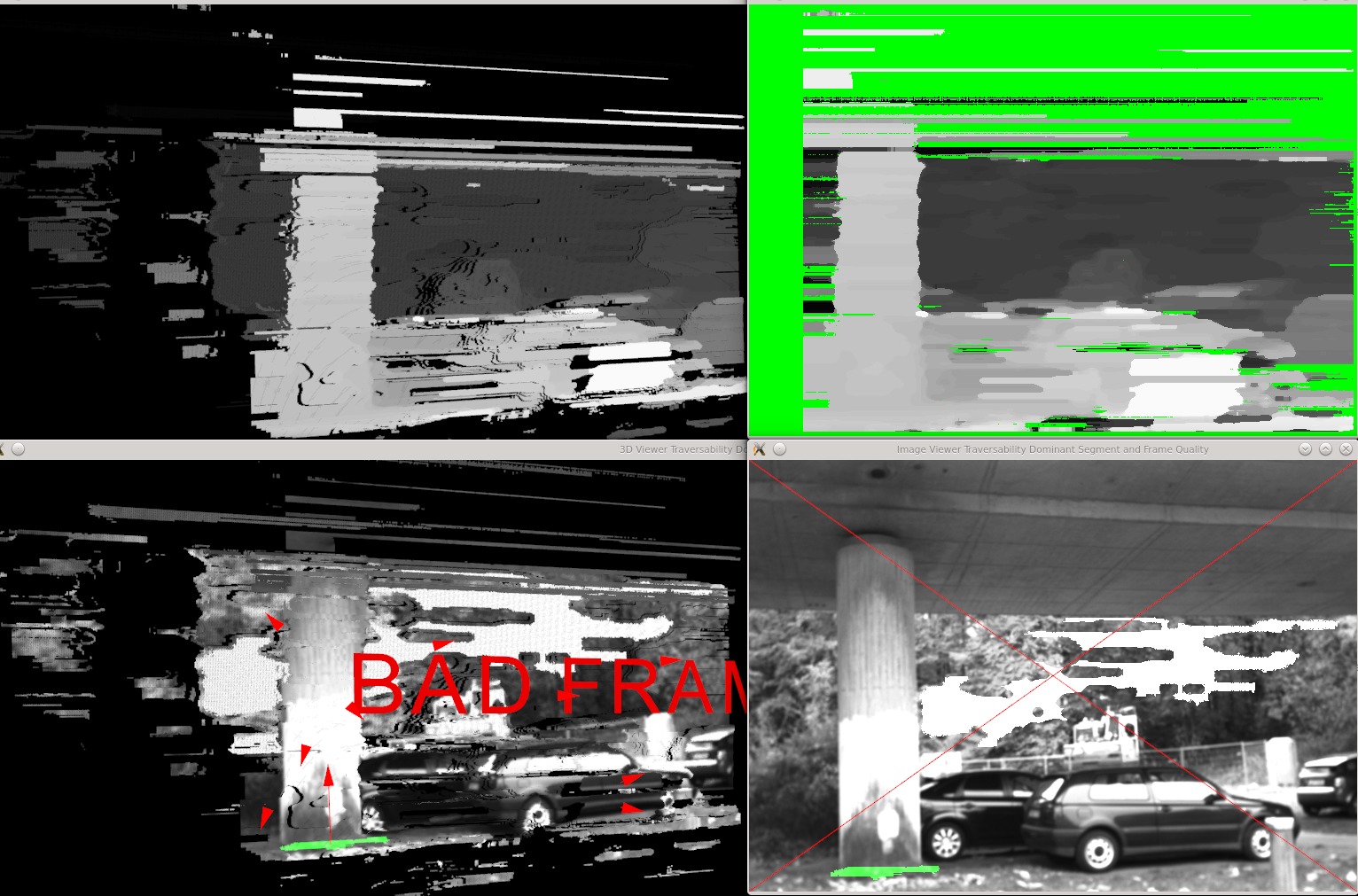}
 \caption{Bad frame no. 12: (top row) the disparity image and (bottom row) the dominant ground plane is shown in white and red indicates the dominant obstacle plane. 
 The arrows on the 3D point cloud in left hand is indicating the detected superpixel surface normals.}
 \label{fig:res:mine:disp_vs_domin12-bad}
\end{figure}

\begin{figure}
 \centering
 \includegraphics[width=1.0\columnwidth]{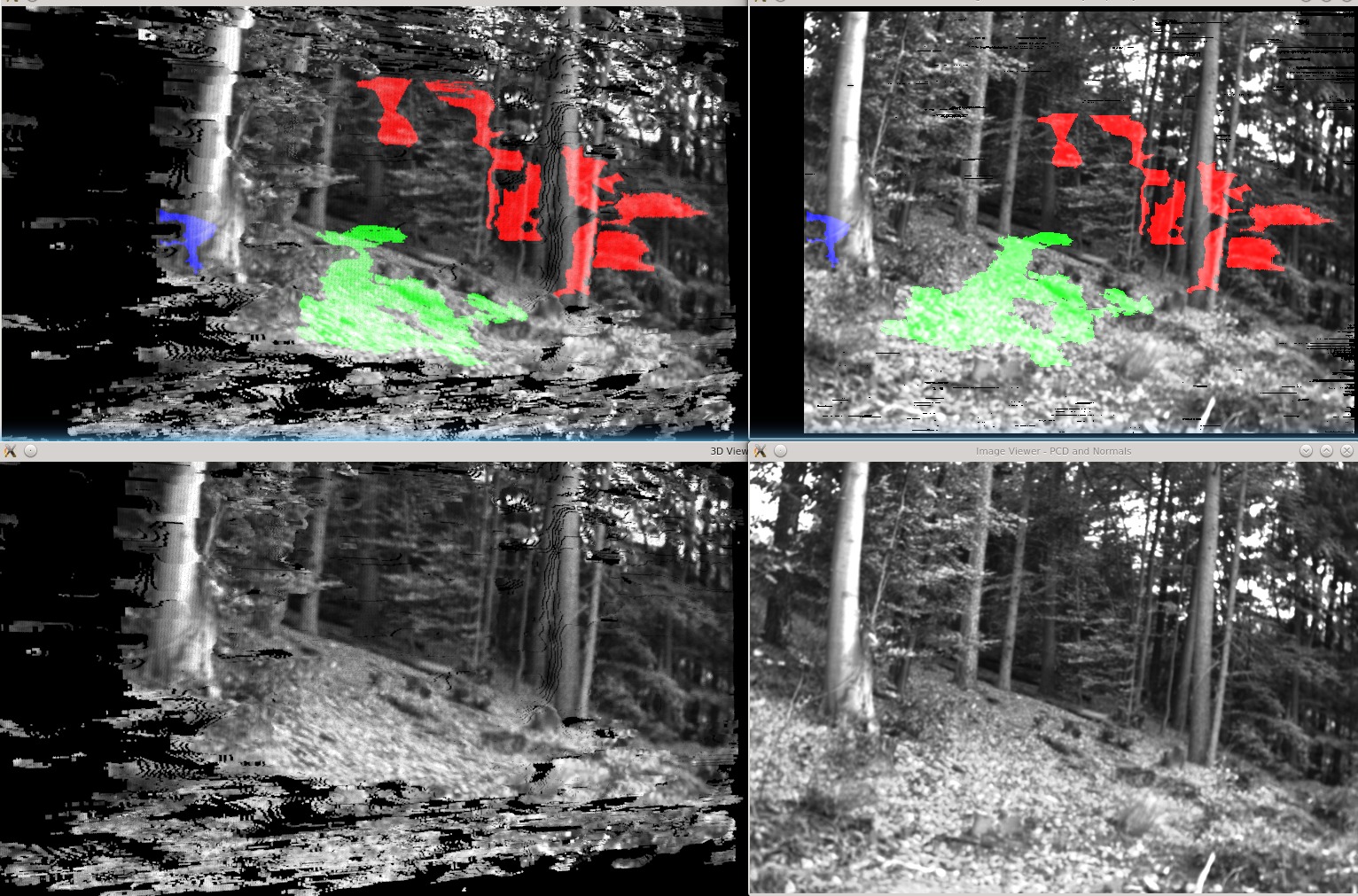}
 \caption{Good frame no. 50: (top row) terrain classification results and (bottom row) 3D reconstructed of an unstructured rough terrain in forest.}
 \label{fig:res:mine:trav_vs_input50}
\end{figure}

A series of gray stereo images are generated by stereo system shown in figure \ref{fig:res:mine:stereo_ravon_gray} along with 
the terrain classification results which are demonstrated in figure \ref{fig:res:mine:trav_vs_input8}, \ref{fig:res:mine:disp_vs_domin9}, 
\ref{fig:res:mine:disp_vs_domin12-bad} and \ref{fig:res:mine:trav_vs_input50}. 

\begin{figure}
 \centering
 \includegraphics[width=1.0\columnwidth]{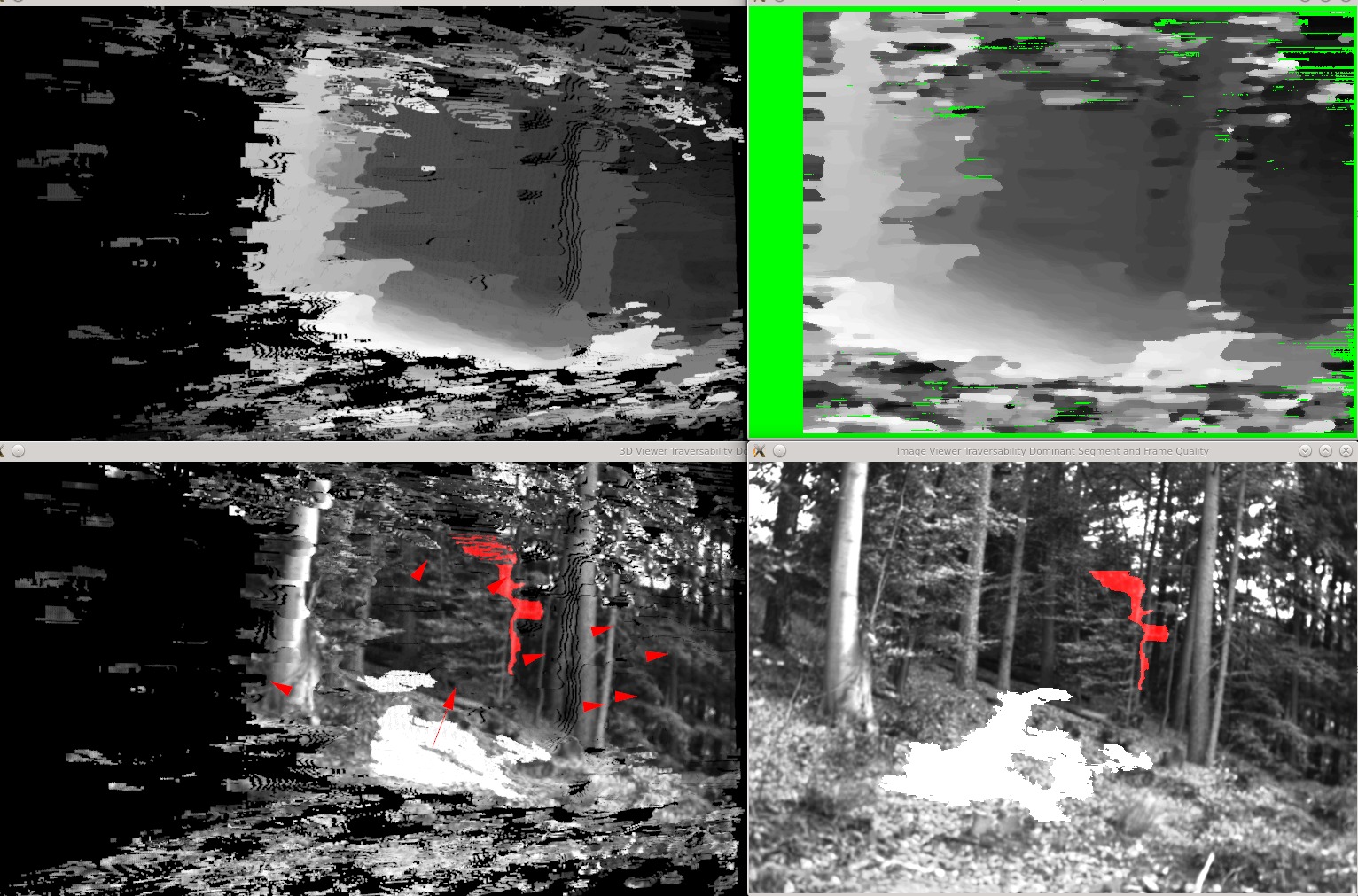}
 \caption{Good frame no. 50: (top row) the disparity image and (bottom row) the dominant ground plane is shown in white and red indicates the dominant obstacle plane. 
 The arrows on the 3D point cloud in left hand is indicating the detected superpixel surface normals.}
 \label{fig:res:mine:disp_vs_domin50}
\end{figure}

\section{Conclusion and future work}
\label{sec:disc}
The proposed approach reformulated the problem of terrain traversability analysis into 
(1) 3D terrain reconstruction and (2) terrain all surfaces detection and analysis.
SSTA approach is proposed for more accurate terrain classification using vehicle kinematics capability.
This approach produces reasonable results in detecting all important surfaces using only geometry-based features such as normals.
That is why some inaccurate surface classification results happen due to noisy generated point cloud.
Using appearance-based features such as textures along with geometry-based ones may help segmentation module 
in detecting less noisy superpixel surfaces.
The classification result can be also used for path planning and passage detection for UGV navigation
since all of the important surfaces in the terrain are detected as superpixel not fixed-size patches.
These are classified based on their traversability content.
That is why pixel-based texture segmentation might increase the accuracy of the surface detection.

\addtolength{\textheight}{-12cm}   




\end{document}